\documentclass[10pt,a4paper,conference]{IEEEtran}
\usepackage{cite}
\usepackage{siunitx}

\ifCLASSINFOpdf
   \usepackage[pdftex]{graphicx}
\else
\fi

\usepackage{booktabs,ragged2e}

\usepackage{url}
\begin{document}
%
% paper title
% Titles are generally capitalized except for words such as a, an, and, as,
% at, but, by, for, in, nor, of, on, or, the, to and up, which are usually
% not capitalized unless they are the first or last word of the title.
% Linebreaks \\ can be used within to get better formatting as desired.
% Do not put math or special symbols in the title.
\title{Enriching Video Captions With Contextual Text}

% author names and affiliations
% use a multiple column layout for up to three different
% affiliations
\author{\IEEEauthorblockN{Philipp Rimle}
\IEEEauthorblockA{ETH Zürich\\
primle@ethz.ch}
\and
\IEEEauthorblockN{Pelin Dogan}
\IEEEauthorblockA{ETH Zürich\\
pelin.dogan@inf.ethz.ch}
\and
\IEEEauthorblockN{Markus Gross}
\IEEEauthorblockA{ETH Zürich\\
grossm@inf.ethz.ch}
}

\newcommand{\TODO}[1]{\textcolor{red}{#1}}
\newcommand{\PAR}[1]{\vskip3pt \noindent{\it #1~}}

% conference papers do not typically use \thanks and this command
% is locked out in conference mode. If really needed, such as for
% the acknowledgment of grants, issue a \IEEEoverridecommandlockouts
% after \documentclass

% for over three affiliations, or if they all won't fit within the width
% of the page, use this alternative format:
%
%\author{\IEEEauthorblockN{Michael Shell\IEEEauthorrefmark{1},
%Homer Simpson\IEEEauthorrefmark{2},
%James Kirk\IEEEauthorrefmark{3},
%Montgomery Scott\IEEEauthorrefmark{3} and
%Eldon Tyrell\IEEEauthorrefmark{4}}
%\IEEEauthorblockA{\IEEEauthorrefmark{1}School of Electrical and Computer Engineering\\
%Georgia Institute of Technology,
%Atlanta, Georgia 30332--0250\\ Email: see http://www.michaelshell.org/contact.html}
%\IEEEauthorblockA{\IEEEauthorrefmark{2}Twentieth Century Fox, Springfield, USA\\
%Email: homer@thesimpsons.com}
%\IEEEauthorblockA{\IEEEauthorrefmark{3}Starfleet Academy, San Francisco, California 96678-2391\\
%Telephone: (800) 555--1212, Fax: (888) 555--1212}
%\IEEEauthorblockA{\IEEEauthorrefmark{4}Tyrell Inc., 123 Replicant Street, Los Angeles, California 90210--4321}}

% use for special paper notices
%\IEEEspecialpapernotice{(Invited Paper)}

% make the title area
\maketitle

% As a general rule, do not put math, special symbols or citations
% in the abstract
\begin{abstract}
Understanding video content and generating caption with context is an important and challenging task. Unlike prior methods that typically attempt to generate generic video captions without context, our architecture contextualizes captioning by infusing extracted information from relevant text data. 
We propose an end-to-end sequence-to-sequence model which generates video captions based on visual input, and mines relevant knowledge such as names and locations from contextual text. In contrast to previous approaches, we
do not preprocess the text further, and let the model directly learn to attend over it. Guided
by the visual input, the model is able to copy words from the contextual text via a pointer-generator
network, allowing to produce more specific video captions. 
We show competitive performance on the News Video Dataset and, through ablation studies, validate the efficacy of contextual video captioning as well as individual design choices in our model architecture.

\end{abstract}

% no keywords

% For peer review papers, you can put extra information on the cover
% page as needed:
% \ifCLASSOPTIONpeerreview
% \begin{center} \bfseries EDICS Category: 3-BBND \end{center}
% \fi
%
% For peerreview papers, this IEEEtran command inserts a page break and
% creates the second title. It will be ignored for other modes.
\IEEEpeerreviewmaketitle

\section{Introduction}
% no \IEEEPARstart

Understanding video content is a substantial task for many vision applications, such as video indexing/navigation~\cite{marques2002content}, human-robot interaction~\cite{zhong2004detecting}, describing movies for the visually impaired people~\cite{rohrbach2015dataset}, or procedure generation for instructional videos~\cite{zhou2018towards}. There are many difficult challenges due to the open domain and diverse set of objects, actions, and scenes that may be present in the video with complex interactions and fine motion details. Furthermore, the required contextual information may not be present in the concerned video section at all, which needs to be extracted from some other sources. 

\begin{figure}
	\centering
	\includegraphics[width=\columnwidth]{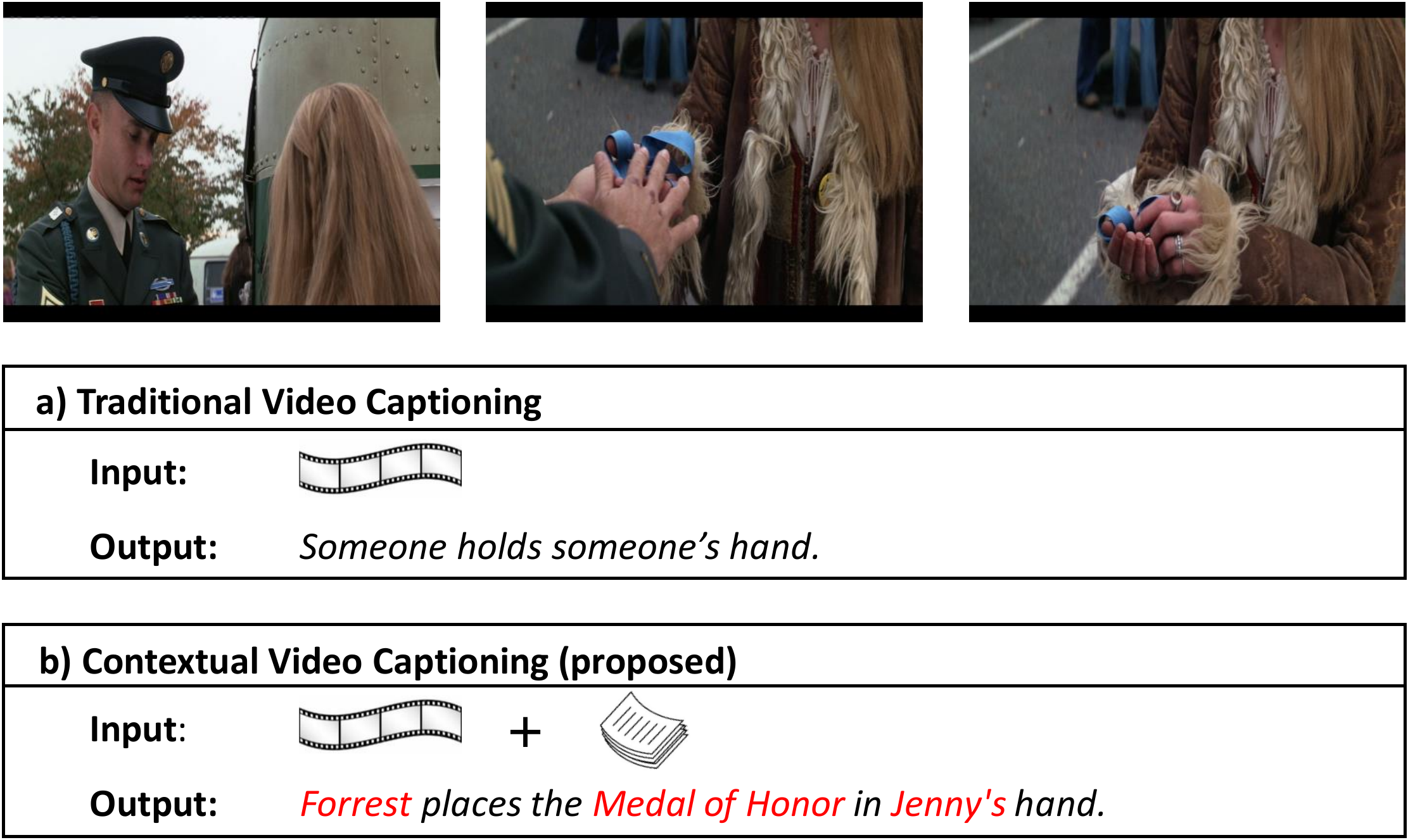}
	\caption{Captions for a video clip from the movie \textit{Forrest Gump}. a) with traditional methods, b) with contextual video captioning that exploits the movie script as an additional input. }
	\label{fig:teaser_forest}
	%\vspace{-7pt}
\end{figure}

While significant progress has been made in video captioning, stemming from release of several benchmark datasets~\cite{chen2011collecting, torabi2015using, guadarrama2013youtube2text, rohrbach2015dataset, lsmdc} and various neural algorithmic designs, the problem is far from being solved. Most, if not all, existing video captioning approaches can be divided into two sequential stages that perform visual encoding and text decoding respectively~\cite{venugopalan15iccv}. These stages can be coupled further by additional transformations~\cite{gao2019hierarchical, wang2018bidirectional} where the models are limited by the input visual content or the vocabulary of a specific dataset. Some approaches~\cite{rohrbach2014coherent} consider the preceding or succeeding video clips to extract contextual relation in the visual content to generate coherent sentences in a storytelling way. In general, these approaches focus on a domain specific dataset not reflecting the whole real world, but only a subset that is missing a lot of information needed to produce human comparable results.
%Several approaches have been proposed in the field of video captioning to overcome the issues. ~\cite{venugopalan15iccv} built an end-to-end sequence-to-sequence model with variable length input and output that is able to learn arbitrary temporal structure in the frame input sequence. The authors of ~\cite{song2018hierarchical} present a hierarchical LSTM with adaptive attention using both spatial and temporal attention for selecting specific regions or frames to predict the related words. ~\cite{wang2018bidirectional} presents a bidirectional proposal method to overcome the challenges of utilizing both past and future video context and feed informative input to the decoder for generating natural event descriptions.
%In general, video captioning models are trained on a specific dataset which does not reflect the whole real world, but only a subset of it missing a lot of information needed to produce human comparable results missing.
Consequently, most captions still tend to be generic like ``someone is talking to someone'' 
%or ``a man is playing guitar on a stage'', 
and the knowledge about \textit{who}, \textit{where} and \textit{when} is missing. We try to overcome this issue by providing contextual knowledge in addition to the video representation. This allows us to produce more specific captions like ``Forrest places the Medal of Honor in Jenny's hand.'' instead of just ``Someone holds someone's hand.'' as illustrated in Figure~\ref{fig:teaser_forest}.

To address these limitations, we propose an end-to-end differentiable neural architecture for contextual video captioning, which exploits the required contextual information from a relevant contextual text. Our model extends a sequence-to-sequence architecture~\cite{venugopalan15iccv} by employing temporal attention in the visual encoder, and a pointer generator network~\cite{see2017point} at the text decoder which allows for extraction of background information from a contextual text input to generate rich contextual captions. The contextual text input can be any text that is relevant to the video up to some degree without strict limitations. This could be a part of the script for a movie section, an article for a news video, or a user manual for a section of an instructional video. 
% \begin{figure}
% 	\centering
% 	\includegraphics[width=\columnwidth,trim={5.3cm 1.6cm 5.8cm 3.4cm},clip]{figures/forest.pdf}
% 	\caption{Captions for a video clip from the movie \textit{Forrest Gump}. a) with traditional methods, b) with contextual video captioning that exploits the movie script as an additional input. }
% 	\label{fig:teaser_forest}
% 	%\vspace{-7pt}
% \end{figure}

\PAR{Contributions.} The contributions of this paper are three-fold. First, we propose a method for contextual video captioning which learns to attend over the context in raw text and generates out of vocabulary words by copying via pointing. The source code for the full framework will be publicly available\footnote{\url{https://github.com/primle/S2VT-Pointer}}.
Second, we augment the LSMDC dataset~\cite{lsmdc} by pairing video sections with the corresponding parts in the movie scripts, and share this new split with the community\footnote{\url{https://github.com/primle/LSMDC-Context}}.
Third, we show competitive performance both with respect to the prior state-of-the-art and ablation variants of our model. Through ablations we validate the efficacy of contextual captioning as well as individual design choices in our model.

\section{Related Work}
Our goal of contextual caption generation is related to multiple topics. We briefly review the most relevant literature below.

\PAR{Unimodal Representations.} It has been observed that deep neural networks such as VGG~\cite{simonyan2014very}, ResNet~\cite{he2016deep}, GoogLeNet~\cite{szegedy2015going} and even automatically learned architectures~\cite{zoph2018learning}, can learn suitable image features to be transferred to various vision tasks~\cite{donahue2014decaf, yosinski2014transferable}. Generic representations for video and text have been receiving considerable attention. Pooling and attention over frame features ~\cite{venugopalan2014translating, xu2015show, yao2015describing}, neural recurrence between frames and spatiotemporal 3D convolution are among the common video encoding techniques~\cite{donahue2015long, ranzato2014video, tran2015learning}. On the language side, distributed word representations~\cite{mikolov2013efficient, pennington2014glove} and recent attention-based architectures~\cite{devlin2018bert, radford2018improving} provide effective and generalisable representations modeling sentential semantics. 

\PAR{Joint Reasoning of Video and Text.} Popular research topics in joint reasoning of image/video and text include video captioning ~\cite{vinyals2015show, xu2015show, pan2016jointly}, retrieval of visual content ~\cite{lin2014visual, anne2017localizing} and text grounding in images/videos~\cite{lin2014visual, fukui2016multimodal, rohrbach2016grounding, plummer2017phrase }. Most approaches along these lines can be classified as belonging to either (i) joint language-visual embeddings or (ii) encoder-decoder architectures. The joint vision-language embeddings facilitate image/video or caption/sentence retrieval by learning to embed images/videos and sentences into the same space~\cite{pan2016jointly, xu2016heterogeneous}. The encoder-decoder architectures [43] are similar, but instead attempt to encode images into the embedding space from which a sentence can be decoded ~\cite{venugopalan2015sequence,yu2016video, gao2019hierarchical}. Most of these approaches yield generic video captions without any context due to lack of background knowledge. 

Contextual video captioning has not received great attention yet besides few attempts~\cite{venugopalan2016improving, chunseong2017attend, tran2017natural} which might be due to lack of suitable datasets. \cite{whitehead2018incorporating} presents a dataset of news videos and captions that are rich in knowledge elements and employs \textit{Knowledgeaware Video Description Network (KaVD)} that incorporates entities from topically related text documents. Similar to~\cite{whitehead2018incorporating}, we incorporate relevant text data, with the use of pointer networks~\cite{see2017point}, for a given video to produce richer and contextual captions. In contrast to KaVD, we propose a model which directly operates on raw contextual text data. Our model learns to attend over the relevant words, based on visual input which allows the model not only to learn contextual entities and events, but also the interaction between them. Further, it allows both video captioning with background knowledge, as well as text summarization based on visual information. We also eliminate the additional preprocessing overhead of name/event discovery and linking systems.

\section{Approach}

\begin{figure*}[!t]
    \centering
    \includegraphics[width=\linewidth]{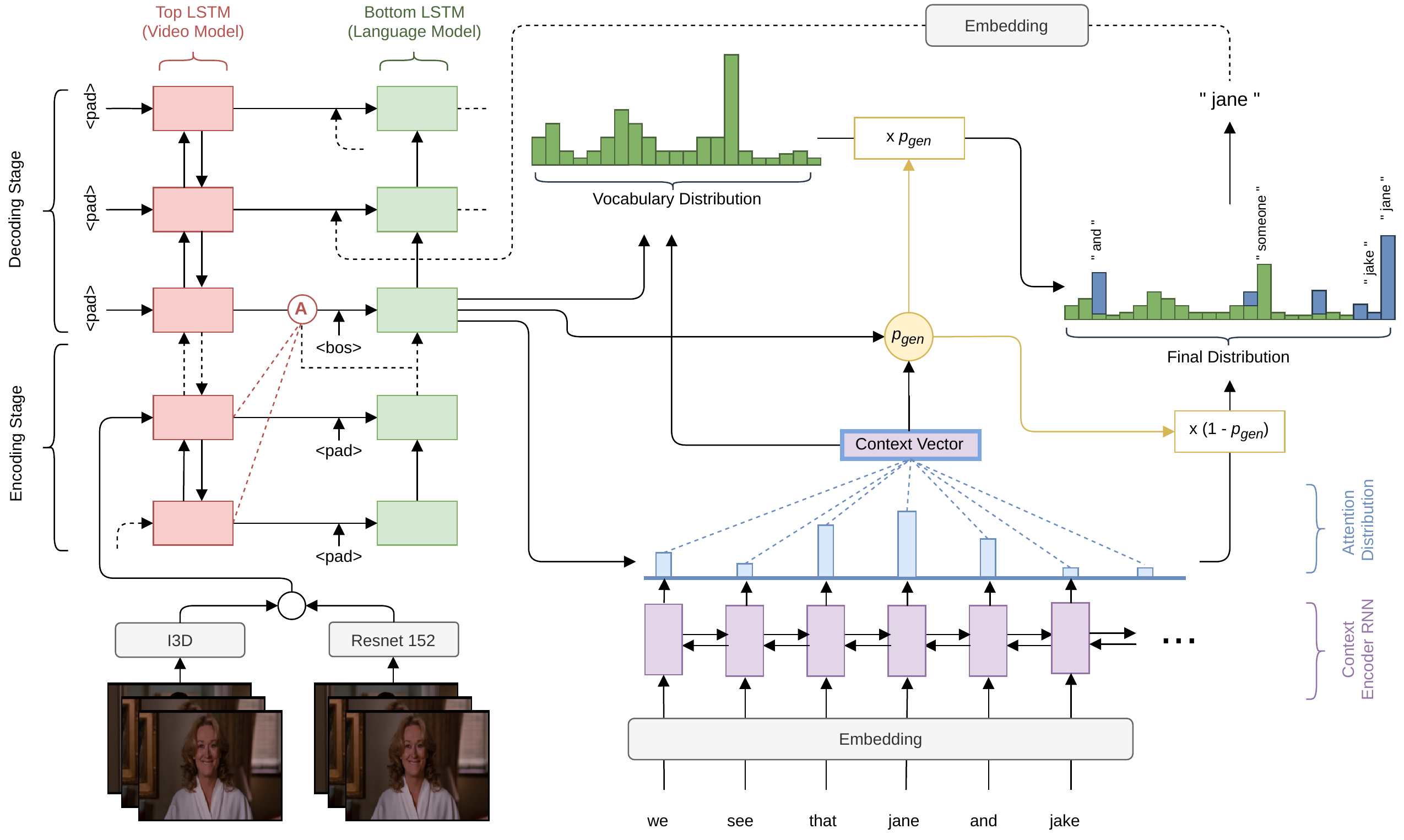}
    \caption{Model Overview. A stack of two LSTM blocks is used for both encoding (red) and decoding (green) the visual input and textual output respectively. The bottom LSTM (green) layer additionally uses a temporal attention to identify the relevant frames. The contextual text input of variable size is encoded using another bidirectional LSTM to build a visual and context aware vocabulary distribution with the use of a pointer generator network.}
    \label{fig_model_overview}
\end{figure*}

We now present our neural architecture for contextual video captioning. An overview of our model is shown in Figure~\ref{fig_model_overview}. 
The input video clip consists of a number of consecutive frames $V = \{V_i\}_{i=1...N}$. The contextual text sequence consists of a number of consecutive words $Z = \{Z_i\}_{i=1...M}$.
Our task is to find a function $\pi$ that encodes the input sequences and decodes a contextual caption as a sequence of consecutive words $Y = \pi(V, Z) = \{Y_i\}_{i=1...L}$. We rely on a sequence-to-sequence architecture to handle variable input and output length. A stack of two LSTM~\cite{hochreiter1997long} blocks as proposed in~\cite{venugopalan15iccv} is used for both encoding and decoding, which allows parameter sharing between the two stages. 
The stack consists of a bidirectional and a unidirectional LSTM which are mainly effective in encoding and decoding, respectively. 
During decoding, the bottom LSTM layer additionally uses a temporal attention over the hidden states of the top LSTM layer to identify the relevant frames. 
Next to the visual input, a contextual text input of variable size is encoded using another bidirectional LSTM. We use a pointer generator network~\cite{see2017point} to attend over the contextual text and build a visual and context aware vocabulary distribution. In addition, the pointer generator network allows us to copy context words directly into the output caption, which enables extracting specific background knowledge not available from only visual input.

\subsection{Encoder-Decoder Network}

The baseline architecture consists of two main blocks: a \mbox{bidirectional} LSTM block stacked on top of a unidirectional LSTM block, modeling the input frame and output word sequences, respectively. The top LSTM takes an embedded video feature vector $v_t$ at time step $t$ as input, and passes its hidden state $s_{t}^{top}$ concatenated with the embedding of the previously predicted word input $x_t$ and a frame context vector $s_{t}^*$ to the bottom LSTM block:
    \begin{IEEEeqnarray}{rCl}
        s_{t}^{top}, c_{t}^{top}        & = & BiLSTM(v_t, s_{t-1}^{top}, c_{t-1}^{top}) \\
        s_{t}^{bottom}, c_{t}^{bottom}  & = & LSTM([x_t, s_{t}^{top}, s_{t}^*], s_{t-1}^{bottom}, c_{t-1}^{bottom}) \quad
    \end{IEEEeqnarray}
where $c_{t}^{top}$, $c_{t}^{bottom}$ are the respectively memory cells for the top and bottom LSTM.

The time axis of the stacked LSTMs can be split into the encoding and decoding stage. During encoding, each video frame is passed into a pretrained Convolutional Neural Network (CNN) to obtain frame level features, from where the linear embedding $v_t$ to a lower dimensional space is learned. Since there is no previously predicted word input and no frame context vector during this stage, a padding vector of zeros is used for $x_t$ and $s_t^*$.
% TODO: maybe mention N and T (e.g. N < T padding up to T frames, N > T cropping to T frames

The decoding stage begins after the fixed amount of encoding time steps $T$, by feeding the \textit{beginning of sentence} (BOS) tag to the model. The BOS tag is used to signal the model to start decoding its latent representation of the video as a sentence. Since there is no video frame input in this stage, a padding vector is passed to the top LSTM. To obtain the frame context vector $s^*_t$ at decoding timestep $t$, a temporal attention with an additive alignment score function \cite{bahdanau2014neural} over the hidden states of the top LSTM $\{s_{0}^{top}, \dots, s_{T}^{top}\}$ is applied:
    % \begin{IEEEeqnarray}{rCl}
    %     \alpha^t_j & = & score(s_{j}^{top}, s_{t-1}^{bottom}) \\
    %     \eta^t_j & = & softmax(\alpha^t_j) \\
    %     s^*_t & = & \sum_j \eta^t_j s_{j}^{top}
    % \end{IEEEeqnarray}
\begin{IEEEeqnarray}{rCl}
        \alpha_{t,j} & = & score(s_{j}^{top}, s_{t-1}^{bottom}) \\
        \eta_{t,j} & = & softmax(\alpha_{t,j}) \\
        s^*_t & = & \sum_j \eta_{t,j} s_{j}^{top}
\end{IEEEeqnarray}

The output of the bottom LSTM $s_{t}^{bottom}$ is then passed to the pointer generator network, generating the output word $y_t$. During the encoding stage, no loss is computed and the output of the LSTM is not passed to the pointer generator network.

%At this stage, there is no previously predicted word. Hence, a padding vector of zeros with the respective word embedding dimension is used. Similarly, we do not attend over the frame representation and as a result the frame context vector $s_{t}^*$ is just a vector of zeros. During the encoding stage, no loss is computed and the output of the LSTM is not passed to the pointer generator network.

\subsection{Pointer Generator Network}

We use a bidirectional LSTM to learn a representation of the contextual text. At each context encoder timestep $i$, the embedded word $z_i$ is passed to the LSTM layer, producing a sequence of context encoder hidden states $h_i$. These hidden states are used to build a soft attention distribution $\xi_t$ over the context word representations per decoder timestep~$t$, similar to~\cite{nallapati2016abstractive}:
    % \begin{IEEEeqnarray}{rCl}
    %     \beta^t_i & = & u^T tanh(W_h h_i + W_s s_t^{bottom} + b_{attn}) \label{equ:score_context_attention}\\
    %     \xi^t_i & = & softmax(\beta^t_i)
    % \end{IEEEeqnarray}
\begin{IEEEeqnarray}{rCl}
    \beta_{t,i} & = & u^T tanh(W_h h_i + W_s s_t^{bottom} + b_{attn}) 
    \label{equ:score_context_attention}\\
    \xi_{t,i} & = & softmax(\beta_{t,i})
\end{IEEEeqnarray}
where $u$, $W_h$, $W_s$ and $b_{attn}$ are learned parameters.
To overcome the general issue of tendency to produce repetition in sequence to sequence models, \cite{tu2016modeling, see2017point} proposed a coverage model, which keeps track of the attention history. At each decoder timestep $t$, we follow the same procedure by introducing a coverage vector $c_t$, which is the sum of the previous attention distributions:
    % \begin{equation}
    %     c^t = \sum^{t-1}_{t'=0} \xi^{t'}
    % \end{equation}
\begin{equation}
    c_t = \sum^{t-1}_{t'=0} \xi_{t'}
\end{equation}
This vector informs the model about the degree of attention that the context words have received so far, and helps the model not to attend over the same words repeatedly. The coverage vector is fed to the pointer-generator network as an additional input, and the attention score calculation from Equation~\ref{equ:score_context_attention} is modified as:
    % \begin{equation}
    % \label{equ:attention_score_coverage}
    %     \beta^t_i = u^T tanh(W_h h_i + W_s s_t^{bottom} + w_c c^t_i + b_{attn})
    % \end{equation}
\begin{equation}
    \label{equ:attention_score_coverage}
    \beta_{t,i} = u^T tanh(W_h h_i + W_s s_t^{bottom} + w_c c_{t,i} + b_{attn})
\end{equation}
where $w_c$ is a learned parameter vector of the same shape as $u$.
The resulting context vector $h^*_t$, computed as
    % \begin{equation}
    %     h^*_t = \sum_i \xi^t_i h_i
    % \end{equation}
\begin{equation}
    h^*_t = \sum_i \xi_{t,i} h_i
\end{equation}
is then concatenated with the decoder hidden state and passed to two fully connected linear layers to produce the vocabulary output distribution $P_{vocab}$:
    \begin{equation}
    \label{equ:p_vocab}
        P_{vocab} = softmax(W'(W[s_t^{bottom}, h^*_t] + b) + b')
    \end{equation}
where $W$, $W'$, $b$ and $b'$ are learned parameters. At each decoder timestep $t$, we additionally calculate a generation probability $p_{gen}$, as proposed in \cite{see2017point}, based on the context vector $h^*_t$, the decoder hidden state $s_t^{bottom}$, and the embedded decoder word input $x_t$:
    \begin{equation}
        p_{gen} = \sigma(w^T_{h^*} h^*_t + w^T_s s_t^{bottom} + w^T_x x_t + b_{ptr})
    \end{equation}
where $\sigma$ is the sigmoid function and the vectors $w_{h^*}$, $w_s$, $w_x$ and the scalar $b_{ptr}$ are learned parameters. The generation probability is used to weight the vocabulary distribution $P_{vocab}$ and the attention distribution $\xi_t$ at timestep $t$. For a word $y$, the final distribution is given as:
    % \begin{equation}
    % \label{equ:final_dist}
    %     P(y) = p_{gen} P_{vocab}(y) + (1-p_{gen}) \sum_{i:z_i=y} \xi^t_i
    % \end{equation}
\begin{equation}
    \label{equ:final_dist}
    P(y) = p_{gen} P_{vocab}(y) + (1-p_{gen}) \sum_{i:z_i=y} \xi_{t,i}
\end{equation}
Note that if a word $y$ is not in the contextual text, $\sum_{i:z_i=y} \xi_{t,i}$ is zero, and similarly if $y$ is not in the global vocabulary, $P_{vocab}(y)$ is zero.

The loss function per decoder timestep $t$ is given as:
    % \begin{equation}
    %     loss = -log P(y_t^*) + \lambda \sum_i min(\xi_i^t, c_i^t)
    % \end{equation}
    \begin{equation}
        loss = -log P(y_t^*) + \lambda \sum_i min(\xi_{t,i}, c_{t,i})
    \end{equation}
where $y_t^*$ is the target word and $\lambda$ is a parameter of the model to weight the additional coverage loss~\cite{see2017point}, used to penalize attending over the same contextual word representation multiple times.

As the coverage mechanism penalizes repeated attention on the contextual text, but not on the global vocabulary, we introduce an additional penalization at inference time. At timestep $t$, the output probability $P(y)$ of a word $y$ is multiplied by the factor $\beta \in [0, 1]$, if it already occurs in the predicted sentence $y_0 \dots y_{t-1}$.

\section{Datasets}

\begin{table*}
    \centering
    \vspace{7pt}    
    \caption[Summary Statistics of the Datasets]{Summary Statistics of the Datasets \label{tab:dataset_stats}}
    
    \begin{tabular}{llrrrrr}
    \toprule
    \textbf{Dataset} & \textbf{Domain} & \textbf{\# Videos} & \textbf{\# Clips} & \textbf{Avg. Duration} & \textbf{\# Sentences} & \textbf{Vocab Size}\\
    \midrule
		MPII-MD~\cite{rohrbach2015dataset}               & Movie             & 94    & 68'337    & 3.9s  & 68'375    & 21'700    \\
		LSMDC~\cite{lsmdc}                          & Movie             & 202   & 118'114   & 4.8s  & 118'081   & 23'442     \\
		News Video~\cite{whitehead2018incorporating}  & News              & --    & 2'883     & 52.5s  & 3'302     & 9'179     \\
		
	\midrule
	
	    LSMDC*                      & Movie             & 177   & 114'039   & 4.1   & 114'039   &  25'204    \\
	    LSMDC-Context-AD            & Movie             & 23   &  14'464  & 4.2  & 14'464   &   8'162    \\
	    LSMDC-Context-Script      & Movie             & 26   & 17'954   &  3.9 & 17'954   &  11'997     \\

    \bottomrule
    \end{tabular}
\end{table*}

We test our approach on two datasets that provide both video and contextual text input. 

\subsection{News Video}

To the best of our knowledge, the News Video Dataset~\cite{whitehead2018incorporating} is the only publicly available dataset consisting both visual and contextual background information for video captioning. The dataset is composed of $2883$ news videos from the AFP YouTube channel\footnote{\url{https://www.youtube.com/user/AFP}} with the given descriptions as ground-truth captions. The videos cover a variety of topics such as protests, attacks, natural disasters and political movements from October, 2015 to November, 2017. Furthermore, the authors retrieved topically related news documents using the video meta-data tags. The official release comes with the video URLs only. However, upon our request, they kindly shared their collected news articles with us, which we use as an contextual text input in our experiments.

\subsection{LSMDC-Context}

The Large Scale Movie Description Challenge (LSMDC) dataset~\cite{lsmdc} is a combination of the MPII-MD~\cite{rohrbach2015dataset} and the M-VAD~\cite{torabi2015using} datasets, consisting of a large set of video clips taken from Hollywood movies with paired audio description (AD) sentences as groundtruth captions. Sentences in the original AD are filtered and manually aligned to the corresponding portions for a better precision by the authors. The released dataset comes with the original captions as well as a pre-processed version, where all the character names were replaced with \textit{someone} or \textit{people}. The latter version is also the most used in related research and benchmarks as the character names come from the movie context rather than the visual input.

To adapt it to our problem, we augmented the LSMDC dataset with additional contextual text by using publicly available movie scripts. The scripts were downloaded from the Internet Movie Script database\footnote{\url{https://www.imsdb.com}},  parsed in a similar way to the public code of Adrien Luxey\footnote{\url{https://github.com/Adrien-Luxey/Da-Fonky-Movie-Script-Parser}}. The extracted text from the scripts were
%We use the indent per row to identify if a certain line belongs to a location name, a stage direction, a characer name or a dialogue. Since the indents differ from script to script, one has to agree on certain spaces in a semi-automatic way at the beginning of the script parsing. In contrast to other approaches,
stored in a location-scene structure and later used to narrow down the contextual text input while generating a caption for a short video clip within the movie. 
Next, we downloaded the movie subtitles\footnote{\url{https://subscene.com/}} and built a coarse \textit{$<$script scene, video time$>$} mapping using the dialogues in scripts. Note that public movie scripts are rare and can be either a draft, a final, or a shooting version. Therefore the stage directions and especially the dialogues may differ from the subtitles a lot. To overcome this issue, we built the mapping in multiple rounds and eliminated the movies and scripts which do not have sufficient correspondences between the video and the script. In the end, we assign a coarse time interval from the movie for each scene in the script which could be used as contextual text input. 

%We first went over the subtitles and checked for exact matches between the subtitle text and the dialogue text of a movie script scene. If a match occurred, we assigned the subtitle's timestamp to the respective movie scene. As a result, every movie script scene has a time interval, either from one or multiple exact subtitle matches or the time interval between two exactly mapped movie script timestamps. In the second round we allowed a weaker matching criteria,
%    \begin{equation}
%    \frac{words_{scene_i} \cap words_{sub_j}}{words_{sub_j}} > p_{weaksearch}
%    \label{equ:weaksearch}
%    \end{equation}
%to assign timestamps to script scenes between exact matches and thus decreasing their time intervals.

\subsubsection{AD-Captions with Context}
Roughly 40 movies from the LSMDC dataset with AD sentences have an available movie script in the form of a draft, a shooting or a final version. In the first step, we analyzed how many words %(not including stop words) 
of the AD-captions can be recovered by the provided movie script context. In the second step, we removed the movies with an average caption/context overlap less than 33.3\% to create a smaller split with better context richness. % maybe remove the rest of this part
This way we can improve the average overlap by 7\% in trade of a smaller but higher quality dataset. The resulting dataset contains 23 movies with a total of $14'464$ video clips. As one expects, experiments have shown that keeping the movies with almost no useful additional context is rather obstructive than helpful in the training process.

\subsubsection{Script-Captions with Context}
A part of LSMDC dataset is composed of movies that are paired with script sentences as groundtruth captions, instead of AD sentences. We used this split as our \textit{toyset} to see how well our model can recover a caption when the ground truth caption is in the contextual text. We select the movies with an available movie script and filter out the movies with a caption/context overlap less than 90\%. The resulting dataset contains 26 movies with a total of $17'954$ video clips.

%Some movies in the LSMDC dataset come with script sentences as reference captions and we are interested in how well our models can recover a caption when the ground truth caption is in the contextual text. We select the movies with an available movie script and filter out the movies with a caption/context overlap less than 90\%. The resulting dataset contains 26 movies with a total of $17'954$ video clips.

\subsubsection{LSMDC*}
The splits above cover a small percentage of the original LSMDC dataset. We denote the bigger split of remaining movies as LSMDC*, % reformulate, since LSMDC* can overlap with the training set of the context splits 
which is to be used to pretrain the encoder-decoder network (without contextual text input) to apply transfer learning for the relatively small splits. The new split contains all video-sentence pairs from the original dataset except the test set, since the groundtruth sentences are not available for the original test set.

%As a result of the hidden (not public available) test set and the overlap with our own LSMDC-Context dataset, a new test, validation and training set of the LSMDC dataset was created. Note that this was necessary to apply transfer learning from the bigger dataset split without context to the smaller splits (AD-caption and script-caption) with context. The new split contains all videos except the test set from the original dataset. 

For these three splits, we created our own training, test, and validation sets considering the number of clips per movie as well as the movie genres. Table~\ref{tab:dataset_stats} shows the statistics of the datasets used.
%We considered the genre as well as the number of clips per video for the generation of our own test, validation and training set.

\section{Experiments}
\subsection{Video and Text Representation}

In all experiments, text data is lower-cased and tokenized into words. For the News Video Dataset, numbers, dates and times are replaced with special tokens following~\cite{whitehead2018incorporating}. A vocabulary is built for each respective dataset and clipped by taking account of the occurrence frequency of words. Each word is mapped to an index and the text input to our model is represented as one-hot vector. Further, we use a pretrained Word2Vec model~\cite{mikolov2013efficient, mikolov2013distributed}, trained on a subset of Google News dataset\footnote{\url{https://code.google.com/archive/p/word2vec}}, to have good initialization to our word embedding layer.

We perform video representation differently depending on the used dataset due to different content and style of videos. 
% TODO: OOV words with temp id (e.g. section text representation master thesis)

\subsubsection{News Video Dataset}
%For each video clip we extract the frames with opencv\footnote{\url{https://pypi.org/project/opencv-python}}. 
For each video clip, we sample one frame per second, as the video clips from the News Video Dataset are longer (up to two minutes) and short-term temporal information is less significant due to the news video style of rapid scene changes.
All frames (RGB images) are smoothed with a Gaussian filter before down-scaling to the size of $224\times224$, to avoid aliasing artifacts. The preprocessed frames are fed into the VGG-16~\cite{simonyan2014deep} pretrained on the ImageNet dataset~\cite{imagenet_cvpr09}, and the output of the second dense layer (fc2-layer, after applying the ReLU non-linearity and before the softmax layer) is fed into the top LSTM of our model. % mentioning LSTM block as well? (instead of just LSTM)

\subsubsection{LSMDC Dataset}
The Large Scale Movie Description Challenge published precomputed video features, which we directly use in all our experiments. They provide two types of features: the output of ResNet-152~\cite{he2015deep} pretrained on ImageNet~\cite{imagenet_cvpr09} before applying softmax, and the output of the I3D model~\cite{carreira2017quo} pretrained on ImageNet and Kinetics~\cite{kay2017kinetics}. I3D makes use of multiple frames and optical flow using 3D CNN, therefore a single feature vector input to our LSTM captures a segment of multiple frames. The concatenation of the two feature vectors is fed into the top LSTM of our model. % mentioning LSTM block as well?

\subsection{Training Setup}

In all our experiments, the video features and text (word) inputs are embedded into a $500$-dimensional and $300$-dimensional space respectively. The LSTMs in the encoder-decoder network have a hidden state size of $512$, and the LSTM block used to encode the contextual text in the pointer generator network has a hidden state size of $256$. During training, dropout~\cite{Srivastava:2014:DSW:2627435.2670313} rate of $0.5$ is applied on the video feature input, embedded word input, embedded context input, and all LSTM outputs. 
The training is performed with the Adam~\cite{kingma2014adam} optimizer using a learning rate of $\num{e-4}$.
% any clipping threshold?

\subsubsection{News Video Dataset}

\begin{figure*}[!t]
    \centering
    \includegraphics[width=\linewidth]{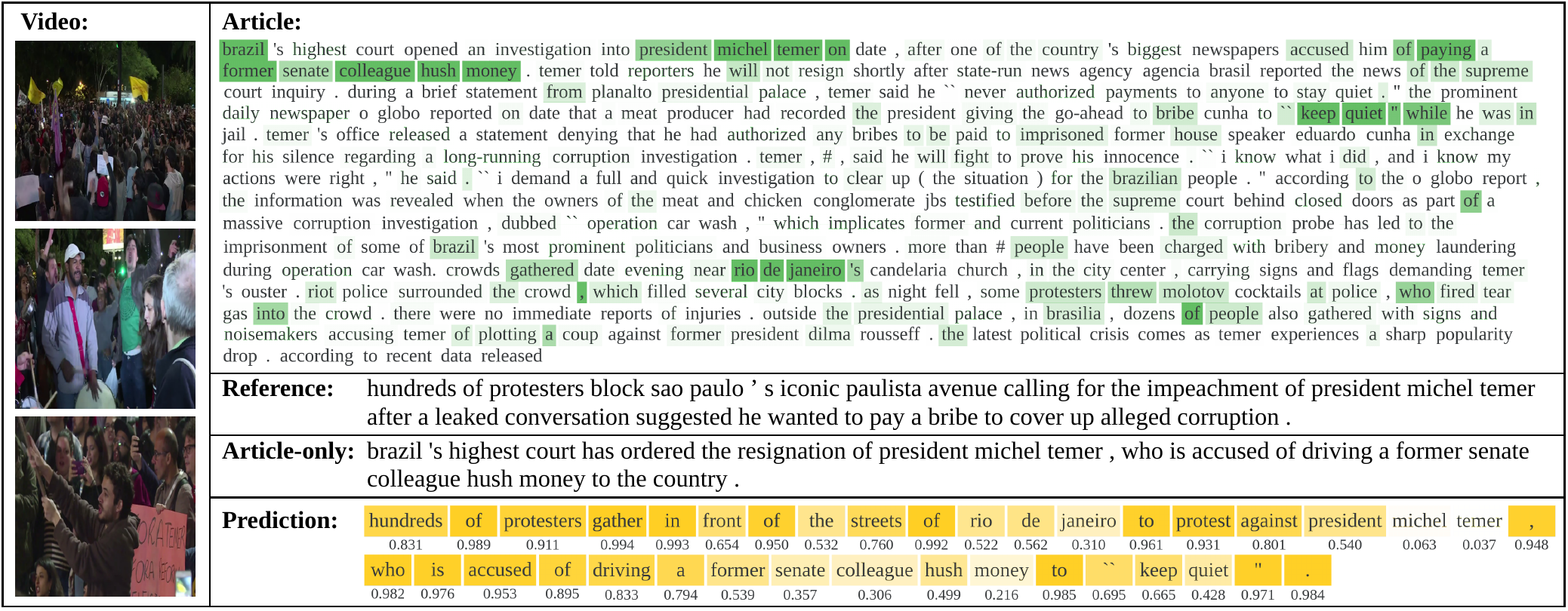} % or *_new.pdf (both are compressed)
    \vspace{-7pt}
    \caption{Sample Prediction on News Video Dataset. Article: the green shading represents the final value of the coverage vector (the sum of the attention distribution for each timestep). A more intense green corresponds with a higher coverage value. Prediction: the yellow shading and the number below represent the generation probability $p_{gen}.$}
    \vspace{-5pt}
    \label{fig_example_result_newsvideo}
    %\vspace{-10pt}
\end{figure*}

We unroll the stacked LSTMs to $120$ timesteps: $60$ for video encoding and $60$ for caption decoding. Note that the News Video Dataset contains longer reference captions than the LSMDC* dataset and mostly includes several sub sentences. Further, we unroll the LSTM for the contextual text to a fixed size of $400$ timesteps, following \cite{see2017point}.
%who found out that increasing the context length decreases their performance significantly. 
Articles are sentence-wise cropped at the end to fit the maximum length of $400$ tokens. For video clips with multiple articles, we create a sample per article and train on all of them. During inference, we take the prediction of the sample/article pair with the highest probability (i.g. most confident). To use transfer learning in some experiments, the complete News Video vocabulary and the most frequent words of the CNN/Daily Mail dataset~\cite{hermann2015teaching, nallapati2016abstractive} were combined together and cropped at $20'000$. We first train the pointer-generator network on the bigger CNN/Daily Mail Dataset, and the sequence-to-sequence model on the News Video Dataset. Secondly, we combine the pretrained models and train on the News Video Dataset. For the final model, we use a coverage loss weight $\lambda$ of $0.2$. At inference time, we use beamsearch with a beamwidth of $2$ and a repetition penalization $\beta$ of $0.2$.

\subsubsection{LSMDC-Context Dataset}

We unroll the stacked LSTMs to $40$ timesteps: $10$ for video encoding and $30$ for caption decoding. Further, we unroll the LSTM for the contextual text to a fixed size of $400$ timesteps for AD-captions and $600$ timesteps for script-captions. Movie script scenes are cropped sentence-wise from the beginning and end, to fit the maximum length of tokens. The complete LSMDC* vocabulary is used for the final models that are trained on LSMDC-Context splits. We first train the sequence-to-sequence model on the bigger LSMDC* dataset with \textit{someone-captions}. Next, we fix the weights of the top LSTM (modeling the video), while training on LSMDC-Context-AD (LSMDC-Context-Script, respectively) with \textit{name-captions}. This procedure provides a good initialization for the pointer-generator network. In a last step, we release all the weights and train the full framework end-to-end. Coverage loss weight $\lambda = 1.0$ is used in the final models. At inference time, we do not use beamsearch (i.e. beamwidth of $1$), but a repetition penalization $\beta$ of $0.2$. 

\subsection{Evaluation}

We use METEOR~\cite{denkowski-lavie-2014-meteor} as our quantitative evaluation metric. It is based on the harmonic mean of unigram precision and recall scores, and considers
how well the predicted and the reference sentences are aligned. METEOR improves the shortcomings of BLEU~\cite{bleu2002} and makes use of semantic matching like stemmed word matches, synonym matches and paraphrase matches next to exact word matches.
%\cite{vedantam2015cider} showed that in evaluating image captions, which is closely related to video captioning, METEOR is always closer to human judgment than BLEU and ROUGE~\cite{lin-2004-rouge}. Further, it outperforms CIDEr~\cite{vedantam2015cider} when the reference set per sample ($1$ in our case) is small. 
In all experiments, we use METEOR 1.5\footnote{\url{http://www.cs.cmu.edu/~alavie/METEOR}} 
%using the code\footnote{\url{https://github.com/tylin/coco-caption}}\footnote{\url{https://github.com/salaniz/pycocoevalcap} (python 3 support)} released with the Microsoft COCO Evaluation Server~\cite{chen2015microsoft}, 
as done in~\cite{venugopalan15iccv}.

\subsection{Results and Analysis}

\begin{table}
    \centering
    \caption[Performance Evaluation on News Video Dataset]{Performance evaluation on the News Video Dataset. \label{tab:score_news_video}}

    \begin{tabular}{lrrrr}
        \toprule
            Model     \hspace{1.5cm}                        &   METEOR [\%]     &   ROUGE-L [\%]    &   CIDEr [\%] \\
        \midrule
            KaVD~\cite{whitehead2018incorporating}   &   10.2            &   \textbf{18.9}            &   --       \\
            
            Video-only                                      &   7.1             &   16.4            &   10.2     \\
            Article-only                                    &   9.3             &   17.2            &   20.5     \\
            S2VT-Pointer                                    &   \textbf{10.8}   &   18.6            &   \textbf{25.7}     \\
        
        \bottomrule
    \end{tabular}
    %\vspace{-5pt}
\end{table}

\begin{figure*}[!t]
    \centering
    \includegraphics[width=\linewidth]{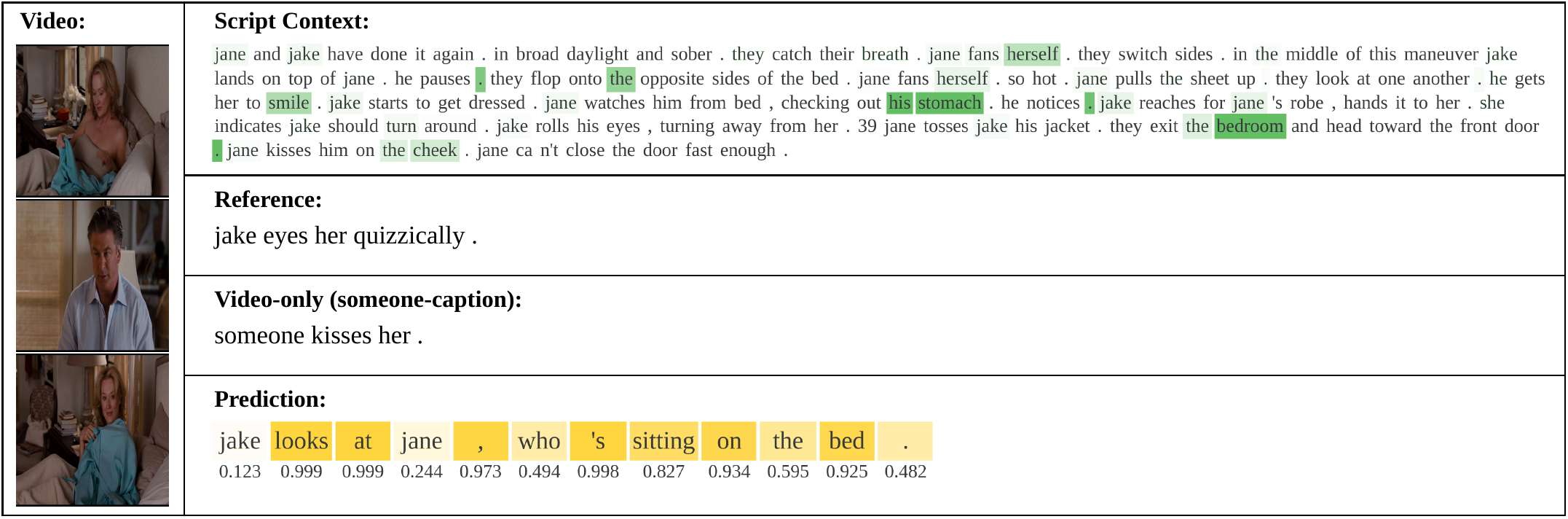}
    \caption{Sample Prediction on LSMDC-Context-AD. Article: the green shading represents the final value of the coverage vector (the sum of the attention distribution for each timestep). A more intense green corresponds with a higher coverage value. Prediction: the yellow shading and the number below represent the generation probability $p_{gen}$}
    \vspace{-7pt}
    \label{fig_example_result_lsmdc}
\end{figure*}

We report the performance of our model on the News Video Dataset in Table~\ref{tab:score_news_video}. In order to understand the benefits of the individual components of our model, we also present an ablation study where
blocks stacks are removed. Our full model performs significantly better than the video-only and the article-only model which are missing the pointer generator network and the video encoder respectively. Comparing the results between KaVD~\cite{whitehead2018incorporating} and our full model is difficult as the authors of KaVD and News Video Dataset only published the ratio of train, validation and test splits, but not the exact sets. The authors did not report the CIDEr score in ~\cite{whitehead2018incorporating}.

We show a qualitative result in Figure~\ref{fig_example_result_newsvideo} to highlight the capabilities of our model which presents a semantically correct summary of the article based on the visual input. While the article focuses on the \textit{hush money investigation}, the model correctly uses this information to augment the visual caption of \textit{protesters doing a demonstration in a street}. This can be seen in the weighting ($p_{gen}$) of the attention distribution and the global vocabulary distribution: words related to the event of \textit{protesting} are taken from the global vocabulary and entities like \textit{rio de janeiro} or \textit{michel temer}, as well as additional information are successfully extracted from the article.

The performance of our model on LSMDC-Context-AD is shown in Table~\ref{tab:score_lsmdc_ad}. The model is able to recover $37.4 \%$ of the character names on average. Figure~\ref{fig_example_result_lsmdc} shows an example where the model correctly extracts the name and scene location from the movie script. The difference between the predicted caption (visually correct) and the groundtruth caption shows the difficulties of the LSMDC dataset in general. Analyzing some example prediction shows that the model occasionally substitutes \textit{someone} with a wrong character name. There are many reasons for this behaviour. Firstly, the movie script context does not necessarily include the video scene, nor the character name. Secondly, the dataset is too small and does not let the model learn a good context model at the pointer generator network. In contrast to experiments on the News Video Dataset that are pretrained on CNN/Daily Mail Dataset, the pointer generator network is missing a good initialization due to lack of larger text corpora with similar content and style for the experiments on LSMDC-Context-AD. 

Table~\ref{tab:score_lsmdc_script} shows the performance on LSMDC-Context-Script.
% , where $90\%$ of the ground-truth caption on average is available in the contextual text input. 
The model is able to learn the mapping between the video and the groundtruth caption that is mostly available in the contextual text. Analyzing some example predictions has shown the issue of the script based captions and why the scores remain relatively low. In LSMDC, consecutive samples tend to have almost identical visual input. Yet, the reference sentences describe different levels of scene details (e.g. \textit{lester, carolyn and jane are eating dinner by candlelight} vs. \textit{red roses are bunched in a vase at the center of the table}). Without the awareness of the sequence of samples, a correct mapping between the script sentences and the reference sentences is ambiguous. This is because a reasonable system would always go for the most likely sentence.

As the ground truth captions from the LSMDC-Context splits highly depend on the respective video clip, we omit the results of the Movie-Script-only model. In contrast to the News Video Dataset, the captions do not reflect a possible summary of the text input and therefore the results are uninformative. 
% This is because a reasonable system would always go for the most likely sentence.

\begin{table}
    \centering
    \caption[Performance Evaluation on LSMDC-Context-AD]{Performance evaluation on the LSMDC-Context-AD dataset. \label{tab:score_lsmdc_ad}}
    \begin{tabular}{lrrrr}
        \toprule
            Model                   & Name-Recovery     &   METEOR          &   ROUGE-L     &   CIDEr  \\
        \midrule
            Video-only              &    --             &  3.4              &  10.1         &  5.2     \\
            S2VT-Pointer            & \textbf{37.4}     &  \textbf{5.8}     &  \textbf{14.0}         &  \textbf{15.3}   \\ % Attention
        \bottomrule
    \end{tabular}
    \vspace{-5pt}
\end{table}

\begin{table}
    \centering
    \caption[Performance Evaluation on LSMDC-Context-Script]{Performance evaluation on the LSMDC-Context-Script dataset. \label{tab:score_lsmdc_script}}
    \begin{tabular}{lrrrr}
        \toprule
            Model               & Name-Recovery     &   METEOR      &   ROUGE-L     &   CIDEr  \\
        \midrule
            Video-only          & --                &  3.5          &  10.1         &  4.7      \\
            S2VT-Pointer        & \textbf{60.0}     & \textbf{13.8} &  \textbf{25.3}         &  \textbf{13.4}   \\
        \bottomrule
    \end{tabular}
    \vspace{-5pt}
\end{table}

\vspace{-5pt}
\section{Conclusion}
In this paper, we proposed an end-to-end trainable contextual video captioning method that can extract relevant contextual information from a supplementary contextual text input. Extending a sequence-to-sequence model with a pointer generator network, our model attends over the relevant background knowledge and copy corresponding vocabulary from the given text input. Results on the News Video Dataset and LSMDC-Context validate the competitive performance of our model which directly operates on the raw contextual text data without the need of additional tools unlike prior methods. Furthermore, we make the source code of our framework and LSMDC-Context publicly available for other researchers. The performance of the presented method is naturally limited by the level of correspondence between the video and the chosen contextual text. In future, we plan to involve multiple contextual resources to extract the relevant contextual information with more confidence and precision.

% conference papers do not normally have an appendix

% use section* for acknowledgment
%\section*{Acknowledgment}

%The authors would like to thank...

% trigger a \newpage just before the given reference
% number - used to balance the columns on the last page
% adjust value as needed - may need to be readjusted if
% the document is modified later
%\IEEEtriggeratref{8}
% The "triggered" command can be changed if desired:
%\IEEEtriggercmd{\enlargethispage{-5in}}

% references section

% can use a bibliography generated by BibTeX as a .bbl file
% BibTeX documentation can be easily obtained at:
% http://mirror.ctan.org/biblio/bibtex/contrib/doc/
% The IEEEtran BibTeX style support page is at:
% http://www.michaelshell.org/tex/ieeetran/bibtex/
%\bibliographystyle{IEEEtran}
% argument is your BibTeX string definitions and bibliography database(s)
%\bibliography{IEEEabrv,../bib/paper}
%
% <OR> manually copy in the resultant .bbl file
% set second argument of \begin to the number of references
% (used to reserve space for the reference number labels box)
\bibliographystyle{IEEEtran}
\bibliography{references}

% that's all folks
\end{document}